# Deep Learning based Quasi-consciousness Training for Robot Intelligent Model

Yuchun Li, Fang Zhang *

School of Chemical and Pharmaceutical Engineering, Changsha University of Science and Technology, Changsha 410114, China

**Abstract**: This paper explores a deep learning based robot intelligent model that renders robots learn and reason for complex tasks. First, by constructing a network of environmental factor matrix to stimulate the learning process of the robot intelligent model, the model parameters must be subjected to coarse & fine tuning to optimize the loss function for minimizing the loss score, meanwhile robot intelligent model can fuse all previously known concepts together to represent things never experienced before, which need robot intelligent model can be generalized extensively. Secondly, in order to progressively develop a robot intelligent model with primary consciousness, every robot must be subjected to at least 1~3 years of special school for training anthropomorphic behaviour patterns to understand and process complex environmental information and make rational decisions. This work explores and delivers the potential application of deep learning-based quasi-consciousness training in the field of robot intelligent model.

**Keywords**: Deep learning; Robot intelligent model; Personality differentiation; Bottom-up paradigms.

## 1. Introduction

Artificial intelligence (AI) has come a long way in the more than 70 years since it was proposed, and AI proceeds on the conjecture that all aspects of learning, or any feature of intelligence, can in principle be accurately described so that a machine can be built to simulate it. Turing argued that in principle computers could mimic aspects of human intelligence [1]. Ray Solomonoff, Founding Father of Algorithmic Information Theory, early proposed priori probabilities assigned sequences symbols on basis lengths for universal Turing machine that required produce interest as output [2]. Although it only began to flourish in the 1990s, machine learning has quickly grown to become the most popular and successful branch of artificial intelligence. Over the last decade, deep learning, as a branch of machine learning, has made major breakthroughs in many tasks that were once thought to be extremely difficult for computers, particularly in the field of machine perception. This field requires the extraction of useful information from images, video, and sound, and given enough training data, deep

* Corresponding author.

E-mail address: zhangf@csust.edu.cn (F. Zhang).

learning is able to extract almost all of the information that a human can extract from perceptual data [1]. To decode nature of biological and create cognitive engine (BrainCog) was proposed [3]. Robotics with Fast Slow Thinking (RFST), a framework that mimics architecture classify makes decisions on based instruction types [4]. L.. Damiano tentatively map characterizing living processes biology produce outline programmatic direction development "organizationally relevant approaches" applying techniques investigative field (embodied) AI [5]. J. Felipe Correa examines the interaction between artificial intelligence (AI), Bayesian statistics, and various key brain structures, such as hippocampus, amygdala, thalamic nuclei, which provide a solid foundation design increasingly advanced human-centric systems [6]. Searching with the keywords robot intelligent model / AI cognition framework / robotic cognitive process, there are 687 documents from 2015-01 to 2025-01, and the average annual number of documents is 69. As shown in Fig. 1a, 2023 reaches the peak of 127 annual publications, and 2019 has the fastest growth rate of 50%, suggesting that the research in this field has been developed rapidly and is in a fast rising stage. From 2015-01 to 2025-01, the top 67 countries/regions in the field of robot intelligent model / AI cognition framework / robotic cognitive process are shown in Fig. 1b, and the countries/regions with the highest number of publications in this field are United The country/region with the most publications in this field is the United States of America (162 publications, 23.58%), and China (108 publications, 15.72%) and United Kingdom (59 publications, 8.59%) are in the second and third places [7].

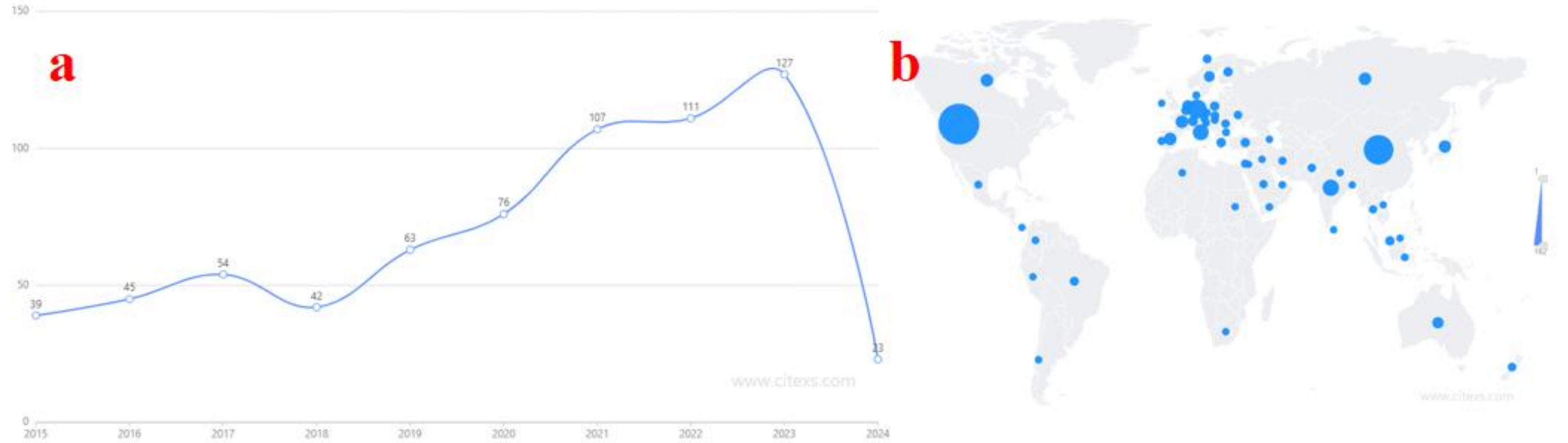

Fig. 1. Bibliographic analysis for annual publication volume and countries of published articles.

From 2015-01 to 2025-01, the top 20 national research institutes in the world in terms of the number of publications in the research field of robot intelligent model / AI cognition framework / robotic cognitive process are shown in Fig. 2a, of which Chinese Academy of Sciences and National Taiwan Normal University occupy the top two places in terms of the number of publications, with 9 and 9 publications, respectively, and Dalian University of Technology has published 7 articles, which is in the third place. Among the 687 papers retrieved, the top 30 journals in terms of the number of publications are shown in Fig. 2b, of which the journal with the most publications is arXiv (Cornell University) (61 papers); Lecture Notes in Computer Science is in the second place, with 12 papers; Advances in intelligent systems and computing with 7 publications [7].

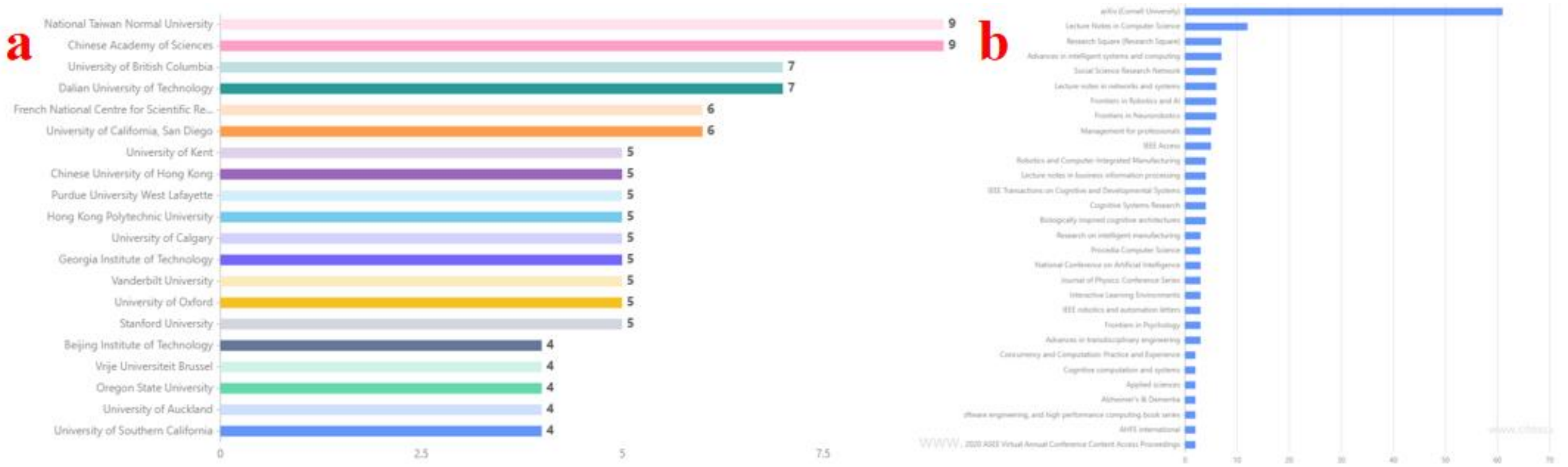

Fig. 2. Bibliographic analysis for ranking of National Research Institutions and Journals.

D. Manishkumar reviewed AI concepts, applications, collaboration models like human-in-the-loop cognitive computing AI. It examines AI's role in improving judgment, handling large datasets, making routine decisions [8]. M. Beer reported the robotic process automation (RPA) in enterprise computing, which is targeted for automating rule-based repetitive and high-volume tasks with higher accuracy which eventually reduces operation cost processing time [9,10]. By adding cognitive artificial intelligence, J. S. Gustaf have studied and concluded that the Robotic process automation (RPA) can be extended, from rule-based, routine processes to more complex applications [11]. All these have seemingly triggered the third AI summer on its own. However, the progress is still far from true robot intelligence. The research on robot intelligent model is relatively not very mature. Existing software infrastructures based on SNNs exclusively support simulation or brain-inspired AI, but not both simultaneously.

Therefore, it is of great practical significance to study deep learning based robot intelligent model for improving robot intelligence level. This paper explored robot intelligent model,which enable robot learn and interact with the environment through continuous consciousness training. In order to progressively develop a robot intelligent model with primary consciousness, personality differentiation training and cultural homogenization training must be carried out at least 1~3 years of special school for robot to promote anthropomorphic behaviour patterns. This work can provide a novel approach for deep learning-based quasi-consciousness training of robot intelligent model.

## 2. Modification of deep learning model structure

As it stands, the label of intelligence in AI is a category mistake; its more accurate name would be artificial cognition or cognitive automation, that's why currently available AI is sometimes considered to be artificial retardation. Actually, AI is a parallel field almost independent of it from each other, in which AI is still a wasteland, with almost all of its content still unexplored. One may find this absurd, isn't something like IBM's Deep Blue very intelligent? Deep Blue implements the intelligence of chess algorithms, not the intelligence of human-like behaviour; because the Alpha-Beta algorithm at the heart of Deep Blue is not a model of the human brain, but just a machine algorithm that only plays chess; it cannot be generalized to any task other than board game-like tasks, and Deep Blue is just a shortcut for researchers to work towards the AI goal of attaining a similar level of intelligence, which has proven to be the case. Deep Blue is like MNIST classification, ImageNet,

Atari Arcade, and even Dota 2, which came after it; they are exceptional in the field of automated cognitive machines, but they are not AIs, and essentially, none of them have any intelligent features. If this statement sounds surprising, remember that human-like intelligence traits are not skills to solve a particular task; it rather is the ability to adapt to new things, to master new skills effectively and to perform never-before-seen tasks [1]. Even if that's the case, T. Giulio pointed out that consciousness can indeed be related to a distributed neural process that is both highly integrated and highly differentiated, and the evidence available so far supports the belief that a scientific explanation of consciousness is becoming increasingly feasible [12]. This is where deep learning needs to be adapted and modified for both large-scale arithmetic simulations and ultra-large-scale model generalization, obviously the latter has a lot of work to do. As shown in Fig. 3, by comparing deep learning based network and actual structure of neurons in the human brain, deep learning based network needs to be transformed towards a diversified, multi-trigger, multi-connectivity model in order to have a strong enough generalization capability.

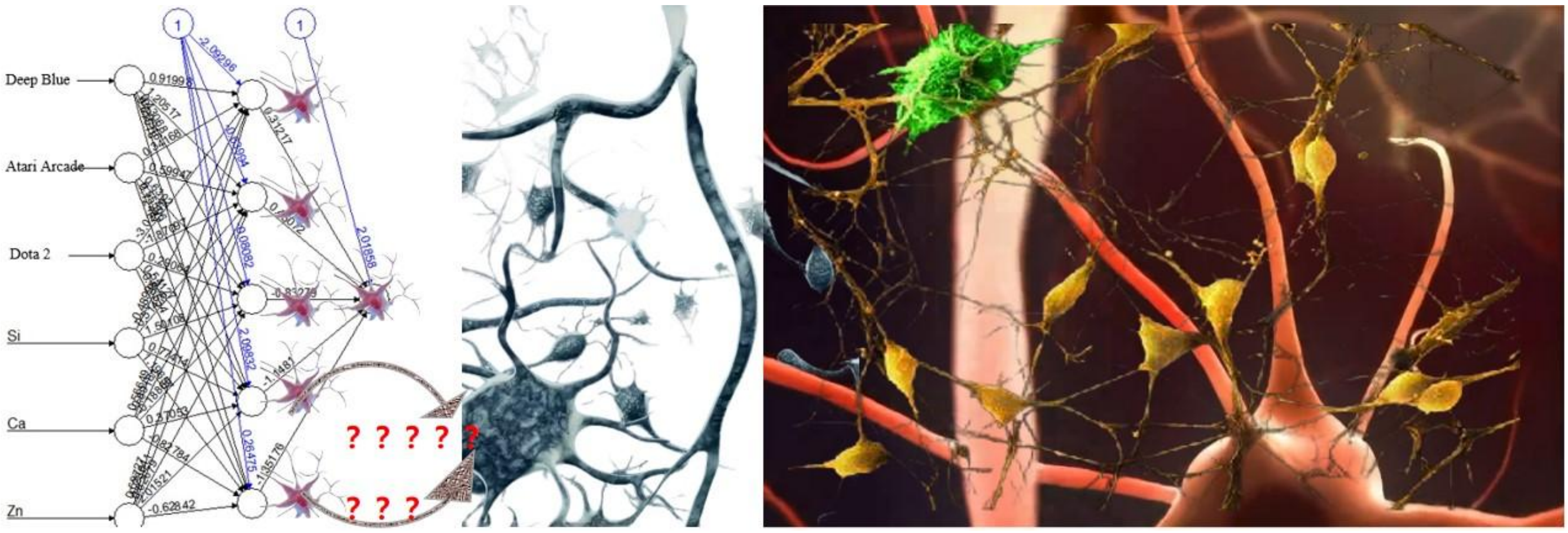

Fig. 3 Comparison between deep learning based network and actual structure of neurons in the human brain.

There is not a single neuron in the human body that is not utilized to its maximum potential, after all, each neuron still needs glial cells to maintain its functioning, whereas our neural network model does not take this into account, the neurons in the model can live in the sky without any consideration of survival, the neuron in the human body cannot be like this, it has to survive and work in the most cost-effective way or it won't survive a single day. So, each real neuron not only receives a large number of receptor stimulation signals, but can be adaptively trained to do multi-task and tuning of multiple receptor models at the same time. By constructing a network of environmental factor matrix to stimulate the learning process of the AI model in robot brain, the model parameters must be subjected to coarse & fine tuning to optimize the loss function for minimizing the loss score.

Deep learning based networks or insects are capable of instantaneous responses to instantaneous stimuli, human beings are capable of much more; we maintain complex abstract models of the current situation, of ourselves, and of other people, and use them to predict various possible futures and to instantly plan for the long term. We can fuse all previously known concepts together to represent things we have never experienced before; this is the ability to abstract and reason, the defining characteristic of human cognitive ability, at which humans are particularly adept. We apply this characteristic to machine learning systems in what is known as extreme generalization; extreme generalization can adapt to new situations that have never been experienced before with very little, if any, new data. This ability is the key to humans having advanced intelligence. Accordingly, the ability of deep networks is limited to local generalization; if the inputs start to deviate from the main trained samples, then the mapping from input to output performed by the deep network becomes immediately meaningless, and the deep network can only generalize to the known unknowns. In this regard, intelligence is characterized by understanding, and the evidence for understanding is the ability to generalize, and the essence of generalization is the ability to deal with a wide range of situations that may arise.

## 3. Robot intelligent model training

### 3.1. Formation of primary consciousness

The robot's mindset should learn and interact with the environment through continuous trial and error, meanwhile continuous thinking training can optimize and improve robot intelligent strategies during the cessation of interaction. However, not all people can be trained to a normal level of intelligence, even if their DNA basis is not any different, which shows that the human intelligence model has two major modules, the physiological basis and the model training basis, which are indispensable for the formation of a model of human intelligence. The process of human mental maturity needs to consider education, family environment, social environment and many other factors, even a drastic event such as a war trauma or a serious assault can completely modify the established model in a person's brain. As long as this point is clarified, the future development of artificial intelligence will have a qualitative indication, which is why most robots must undergo 1~3 years of 'rich and real' environmental training to become real intelligent robot with 'initial consciousness breakthrough'. Just as the process of educating and nurturing people (call it training for robots), whether it is a deliberate process or one that occurs by chance, can also produce psychopaths, fools, or retards, etc., so we must likewise allow for a small probability that robots will have a crippling effect after what we call their mental training, and that they won't be allowed to enter society in the service of mankind, but of course at a much lesser cost relative to people.

As a person grows up, numerous response models are formed in the brain, and each model may also be changed or adjusted after facing some stress, as shown in Fig. 4a, each ball corresponds to a response model, and the large size of the model ball represents the consumption of more resources. When stimulated by a stressful stimulus, the corresponding model ball produces a response, firstly, the stimulus information is transferred to the internal model ball, the internal structure of the model ball is shown in Fig. 4b, the neuron receives the stimulus through the dendrites, and the model ball produces the output, which is then transferred to the relevant mechanism to produce a response. Table 1 provides partial relation structure of neurons, which summarize four kinds of response mode. The first type of neuron response belongs to instinct feedback, which is a one-to-one relationship. The second type is a one-to-many relationship, which may response different way in different situation. The third type is the area in which technology currently excels the most, requiring acquired training to learn for humans, and can be omitted altogether for robots engineering. The fourth type is the one that robots are currently least good at, and the area where deep learning needs to break through the most. In situations where innovation, extreme stress and fantasy are required, AI, like a human being must be able to keep generalizing, sifting, until it finally gives a model to make a decision.

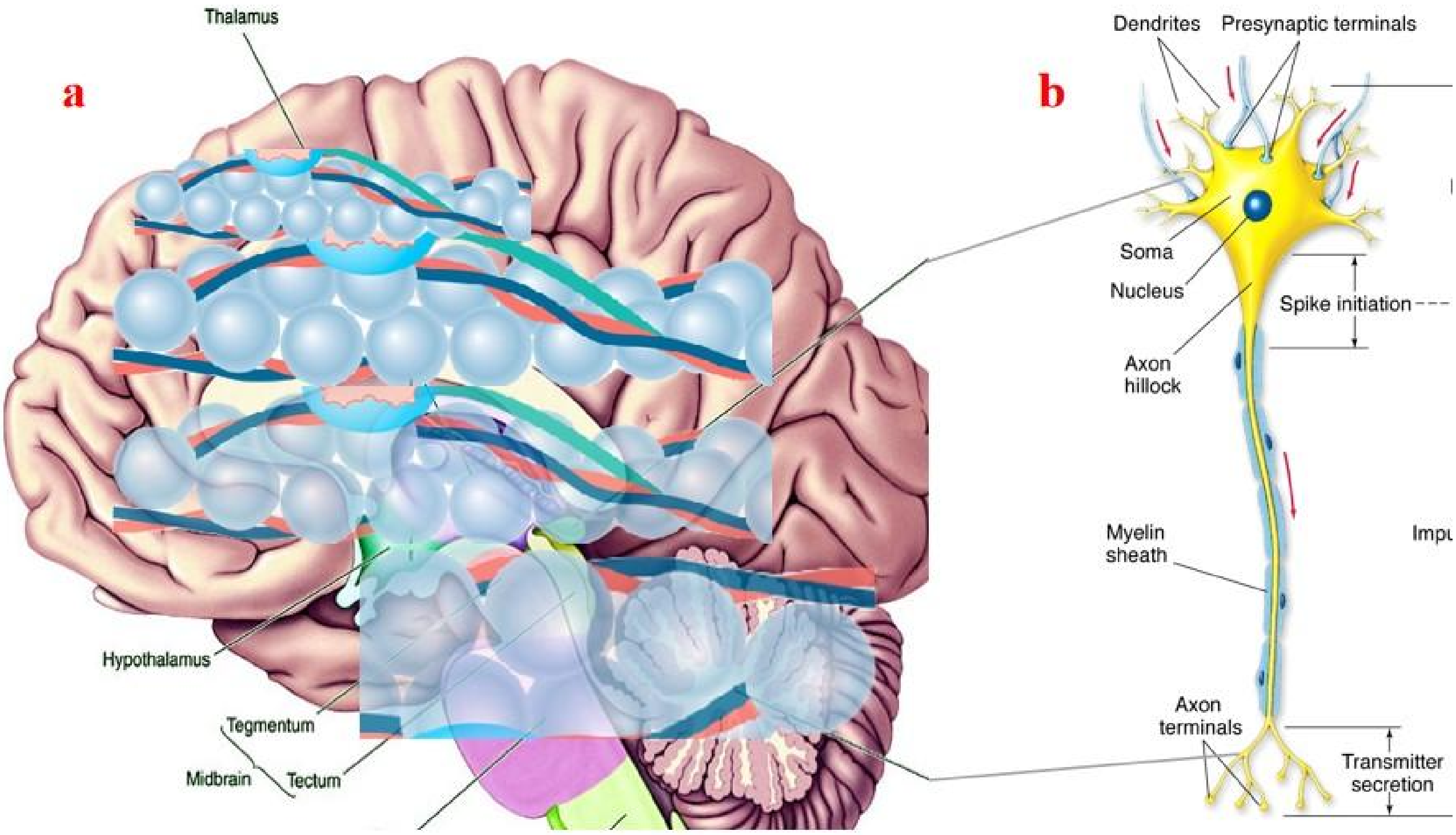

Fig. 4. Schematic of the response model of the human brain.

Table 1.

Relation structure of neuron response.

| No. | Response mode of neurons | Description and application occasions |
|---|---|---|
| 1 | Fixed response mode, inborn | Instinct response, peripheral nervous system (CNS) |
| 2 | Loose response mode, acquired | Perceptual stimulus, tend to be humanities and social sciences feeling emotional |
| 3 | Fixed response mode, acquired | Conclusive science & technology systems |
| 4 | Optimization response mode, acquired & modified deep learning | Innovative, extreme stress, fantasy category |

The first type of neuron response corresponds to general animal level; The second to fourth type correspond to cases of intelligent consciousness.

3.2. Bottom-up design strategy

The essential feature that distinguishes human beings from animals, and which is fundamental to consciousness, is verbal communication and adeptness in the use of tools. In principle, an AI can be called truly intelligent as long as it has the ability to use a variety of tools to the fullest and to communicate verbally to achieve results. AI design strategy for top-down paradigms can be suitable for the first and third response mode of neurons in Table 1. However, the second and fourth response mode of neurons need bottom-up paradigms. Bottom-up paradigms for robot do not provide off-the-shelf model architectures, instead it provides blank model space waiting to be filled in by models generated during the training process.

Bottom-up paradigms can be realized through diversity design and training. Robot diversity design includes differentiation of robot types, which means different inherent style of behaviour. Diversity training for robot indicates special training school for consciousness fostering, which provides training of cultural homogenization and personality differentiation. The models produced by the training school need to be obtained through robot sleep time, so anthropomorphic standby sleep is a stage in the formation of robot intelligence. In addition to the power shortage crisis, the training school needs to provide a variety of crisis environment stimuli to reinforce the formation of awareness in the decision-making process of the robot in the face of a crisis. Special training school for robot is allowed to produce groups of robots with varying proportions of intelligence levels within 1 to 3 years, as shown in Table 2.

Table 2.

Intelligence levels of robot by special training school.

| No. | Intelligence levels | Description |
|---|---|---|
| 1 | Not clever | Intelligence level of human retardation |
| 2 | Not silly | Lower human intelligence level |
| 3 | Normal | Low to medium human intelligence |
| 4 | Clever | Medium level of human intelligence |

## 4. Conclusions

This paper explored robot intelligent model,which enable robot learn and interact with the environment through continuous consciousness training. By constructing a network of environmental factor matrix to stimulate the learning process of the robot intelligent model, the model parameters must be subjected to coarse & fine tuning to optimize the loss function for minimizing the loss score. In order to progressively develop a robot intelligent model with primary consciousness, every robot must be subjected to at least 1~3 years of special school for training anthropomorphic behaviour patterns. This research provides potential applications for deep learning-based quasi-consciousness training in the field of robot intelligent model.

## Acknowledgments

The authors thank the support from Hunan Provincial Key Laboratory of Cytochemistry (Changsha University of Science & Technology), Hunan Provincial Key Laboratory of Materials Protection for Electric Power and Transportation (Changsha University of Science & Technology).

## Declaration of Competing Interest

The authors declare that they have no known competing financial interests or personal relationships that could have appeared to influence the work reported in this paper.

## Data availability

Data will be made available on request.